Machine learning methods fail to provide cohesive atheoretical construction of personality traits from semantic embeddings

Authors:

Ayoub Bouguettaya[1], Elizabeth Stuart[2]

1. Monash University, School of Nursing and Midwifery, Melbourne, Victoria, Australia
2. Miller Children's Hospital, Neuropsychology, Long Beach, California, USA

Abstract

The lexical hypothesis, which suggests that individual differences are coded into language, is the foundation of modern personality psychology. Here, we test this hypothesis using novel machine learning methods to create a bottom-up, atheoretical model of personality from the same trait-descriptive adjective list that led to the dominant, contemporary model of personality (the Big Five). We then compare the descriptive utility of this machine learning method (resulting in lexical clusters) by comparing it to the established Big Five personality model in how well these describe conversations online (on Reddit forums). Our analysis of 1 million online comments shows that the Big Five model provides a much more powerful and interpretable description of these communities and the differences between them. Specifically, the dimensions of Agreeableness, Conscientiousness, and Neuroticism effectively distinguish Reddit communities. In contrast, our lexical clusters do not provide meaningful distinctions and fail to describe the spread. Validation against the International Personality Item Pool confirmed the Big Five model's superior psychometric coherence, and our machine learning methods notably failed to recover the trait of Extraversion. These results affirm the robustness of the Big Five, while also showing that the semantic structure of personality is likely depending on social context. Our findings suggest that while machine learning can help with understanding and explaining human behavior, especially by checking ecological validity of existing theories, machine learning methods may not be able to replace established psychological theories.

**Main**

The advancement of machine learning and large language models (LLMs) presents a fundamental challenge to scientific fields built on human-derived theoretical models. It raises the question of whether foundational theories, derived from decades of subjective interpretation, can be objectively reproduced or even improved by data-driven methods. The study of personality traits, defined as enduring patterns of thoughts, feelings, and behaviors, is a central focus of behavioral science research (Allport, 1927) that may be improved through the use of machine learning methods. Personality psychology was founded on the lexical hypothesis—the idea that significant individual differences are encoded in language (Goldberg, 1993). This lexical hypothesis is the basis for modern personality theory, but was established through methods that machine learning can now re-examine with objective, replicable precision. Therefore, we sought to test machine learning methods to develop better models of personality, and compare them to existing models in how well they describe real world conversations, using Reddit data.

The lexical hypothesis has had significant impacts on personality psychology models (Goldberg, 1993). This principle is the main reason for the development of models like the Big Five, but these models were developed through manual, subjective sorting of trait words—a process that may have embedded researcher bias. Early work by Allport and Odbert (Allport & Odbert, 1936), and later Cattell (Cattell, 1996; Cattell, 1945), exemplified this method. After all, Allport and Odbert first manually examined over 18,000 words to remove perceived redundancies, and then using local raters to sort the remaining terms, which may have solidified this bias. These local raters—mostly in the United States—sorted them based on their own perspectives of semantic overlap. Later, Cattell continued this work through examining semantic overlap statistically, through the newly developed factor analysis method.

This line of research culminated in a five-factor model, and later, the Big Five (Cattell, 1996; McCrae & Costa Jr, 1997), which remains the dominant framework in personality psychology. Modern personality psychology suggests this cluster of personality traits (especially in the form of Costa and McRae's "Big Five" (McCrae & Costa Jr, 1997)) has applications in personality (Costa & McCrae, 2008), clinical (Costa & McCrae, 1992), occupational (Costa Jr, 1996), and forensic psychology (Decuyper et al., 2008). However, despite its predictive validity, the Big Five's universality has been questioned, particularly its stability across non-Western cultures (Laajaj et al., 2019). This further reinforces the possibility that the model's structure is heavily influenced by the subjective judgments of its originators and the raters, who were culture bound (Block, 2010).

Recent advances in computational linguistics, natural language processing, and machine learning have offered new ways for re-examining these established psychological constructs. Research using semantic embeddings (derived from language models) has demonstrated the potential to analyze large datasets of psychometric items, showing the empirical relationships between measures and identifying potential redundancies (Cutler & Condon, 2023; van der Linden et al., 2025). This approach is promising, as concept representations that mimic human understanding appear to emerge naturally in language models (Du et al., 2025; Gigerenzer, 2024; Pellert et al., 2024), suggesting they can, in theory, help organize and improve psychological research, similar to how human raters were used to group words together (Allport & Odbert, 1936). For example, work by Wulff and Mata (Wulff, 2025; Wulff & Mata, 2025) has shown it is possible to use these methods to re-examine existing psychometric scales, refining them to focus on a single construct at a time, and address overlaps between scales. However, while this line of research is critical for improving existing measures, much of this research still operates on personality scales that are products of the original, subjective sorting methods.

Other work has returned to the source of the lexical hypothesis itself, bypassing these potentially flawed scales (Cutler & Condon, 2023). This "deep lexical hypothesis" research applied natural language processing (NLP) directly to the original trait-descriptive adjective lists, confirming that while three factors of personality from the Big Five (Conscientiousness, Extraversion, and Agreeableness) are robustly recovered from language, the factors of Neuroticism and Openness were not. Building on this research, even more recent research used LLMs to examine a large textual internet corpus, and found a large, dominant General Factor of Personality at the top of the personality hierarchy. This primary factor represented a continuum of social desirability and was nearly identical to the general "Evaluation" factor identified in the original human-rated lexical studies (van der Linden et al., 2025). The discovery of a robust, overarching evaluative factor using a purely computational approach further complicates the traditional five-factor view. This leaves a critical question unanswered: does this atheoretical, data-driven structure provide a meaningful and powerful description of personality in naturalistic social contexts?

Here, we address this question by re-examining personality using the lexical hypothesis and relying on the semantic foundations of trait-descriptive language. Our work aims to assess potential flaws in Big Five research by applying a multi-stage computational study to test the structure and utility of personality language in a complex, ecologically valid digital environment. First, to create a new, high-level description of personality, we construct a purely lexical model by applying K-Means clustering to the semantic embeddings of 2,818 trait-descriptive adjectives compiled from foundational psycho-lexical research (Condon et al., 2022). Next, we build a contextual model by analyzing the embeddings of these adjectives as they appear in a corpus of one million comments from Reddit. We then compare the descriptive power of these two bottom-up models against the established Big Five framework to differentiate the linguistic profiles of these online communities. Finally, to provide a rigorous test of the applicability of these models on personality scales, we validate all three models by assessing their semantic alignment with items from the full International Personality Item Pool (IPIP) (Goldberg & Saucier, 2008; Goldberg, 1982, 1992; Goldberg, 1993).

**Results**

*Atheoretical clustering of trait adjectives of structure-based groupings: the "Bottom-up" Lexical Model.*

We first sought to establish an atheoretical structure of personality language by applying K-Means clustering to the scaled semantic embeddings of 2,818 trait-descriptive adjectives via a sentence-embedding transformer model. This initial list was extracted from the original lists used by personality researchers to create the Big Five (Condon et al., 2022). Trial and error methods resulted in producing six distinct semantic clusters. Figure 1 demonstrates the clustering of these adjectives. The analysis revealed that the internal coherence of these clusters, measured by the mean cosine similarity of member words to their centroid, was strongly associated with their linguistic structure. The most cohesive group (cohesion score = 0.70) was a cluster defined by adjectives describing social and personal inadequacy (representative words: 'unlovable', 'unattractive', 'incompetent', 'unintelligent', 'unlikable'). This initial result suggests that, in the absence of social context, the semantic space of trait language is organized by a negative dimension, rather than a multifaceted personality structure. This finding provides only weak support for the lexical hypothesis, as the structure seems to have been driven by basic word features rather than the psychological concepts expected.

Figure 1: Semantic clustering of Personality Clusters for 2,818 trait words

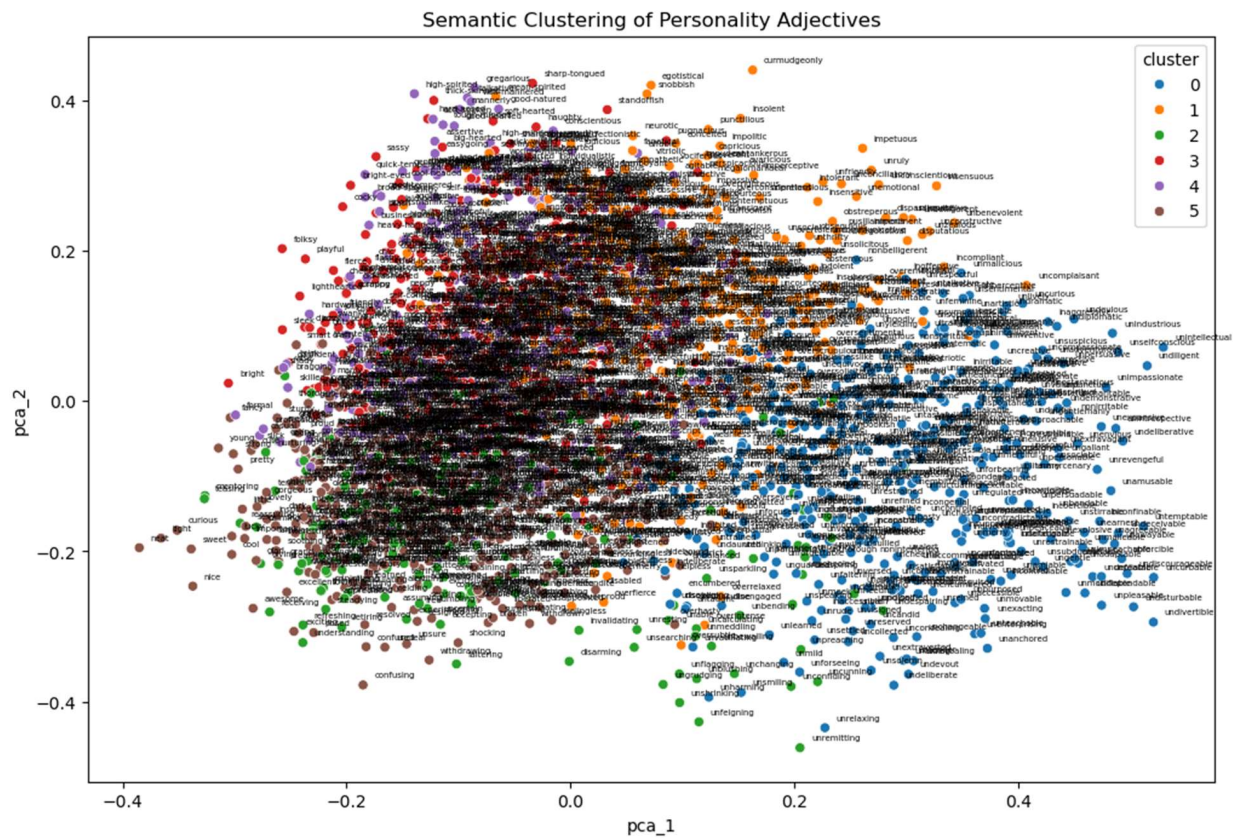

*Adjectives in context of online community use: The "Bottom-up" Contextual Reddit Model*

To ground our analysis in a more naturalistic context, we next identified 1,945 of 2,818 adjectives within a one-million-comment corpus from Reddit (Demarco, 2023). Clustering the embeddings of these context-specific words produced a simplified six-cluster model of personality language. This Contextual Reddit Model revealed a linguistic landscape dominated by two primary functions. A "general description" cluster (representative words: 'effective', 'persistent', 'patient') accounted for most adjective use across the top subreddits (62–75%). The second most prevalent cluster represented a "negative affect" dimension (representative words: 'upset', 'angry', 'unhappy'), which was most prominent in communities focused on interpersonal conflict, such as r/relationshipadvice (28.9% of adjective use). While informative of Reddit communities, this result suggests that machine learning approaches, even when examining embeddings in context, fail to have clear clustering in ways that would support the lexical hypothesis. In other words, while these clusters describe how subreddits work, they do not describe personality particularly well.

*Describing online communities: the Big Five, the Lexical Model, and Contextual Reddit Model*

To test which framework best describes the language of online communities, we compared the descriptive power of our two data-driven models against a machine learning derived Big Five model. We created this top down Big Five model by identifying 50 adjectives commonly used in Big Five scales via the IPIP (Goldberg, 1992).

The analysis showed that the Big Five Model provided specific and interpretable differentiations between subreddits, while the Initial Lexical Model performed poorly. The results of the contextual Reddit model did have some benefits, but was still outperformed by the Big Five model. The Big Five Model successfully described and distinguished the linguistic profiles of the communities. For example, it identified r/ADHD as having the highest use of Conscientiousness-related language (15.5%), while r/marriage had the highest use of Neuroticism-related language (19.11%). Across all communities, Agreeableness was the most prevalent trait category (47.3–58.7%), while Extraversion was surprisingly scarce (less than 11%).

By contrast, both bottom-up models performed poorly in describing the Reddit comments. The Initial Lexical Model, derived from context free words, largely failed to describe and distinguish them meaningfully. This was also unable to detect variance between communities; its distribution was overwhelmingly dominated by a single "negative affect" cluster (Cluster 5), which accounted for over 83% of all adjective mentions in most subreddits. The Contextual Reddit Model, while offering a more nuanced description of subreddits, also showed a highly skewed distribution. It described the communities as being composed primarily of a "negative/affective" cluster (Cluster 1, at around 64% prevalence) and a "general competence" cluster (Cluster 0, around 26% prevalence), offering less descriptive detail than the Big Five framework. These results demonstrate that the top-down, theory-driven Big Five framework provides a more powerful and interpretable description of language in these online communities than either of our bottom-up, data-driven approaches, further suggesting that the lexical hypothesis is not sustained through these machine learning methods.

*Validation: the 300 IPIP NEO items*

To validate our models against established psychometric constructs, we performed a three-stage analysis using the 300 items of the IPIP-NEO personality inventory (Goldberg & Saucier, 2008). These validation steps confirmed that our data-driven clusters captured psychologically meaningful concepts, and reinforced the finding that the Big Five framework provides the most robust (and descriptively superior) structure.

First, we validated that our Contextual Reddit Model clusters mapped onto meaningful psychological dimensions. We did so by identifying the IPIP items most semantically similar to each of the six Reddit cluster centroids. For example, we were able to demonstrate that our fourth cluster (which was characterized by "antagonistic" words) mapped onto IPIP items that described negative social behaviors (like "tease people" or "interrupts others"). Meanwhile, the analysis revealed coherent mappings between our data-driven clusters and established psychological concepts. For example, the Reddit "antagonism" cluster (Cluster 4) mapped to IPIP items describing negative social behaviors, such as "Tease people" and "Interrupt others". The Reddit "negative affect" cluster (Cluster 1) mapped to items related to low self-esteem, such as "Am inadequate" and "Am ashamed of myself". This confirms that the clusters derived from online discussions correspond to meaningful psychological dimensions at least to some degree. These clusters are visible in Supplemental Data code.

Next, we conducted a direct numerical comparison to determine which of our three models best categorized the 300 IPIP items. The "fit score" was calculated as the average similarity of each IPIP item to its closest concept within each model. The Big Five Model achieved the highest fit score (0.3121), outperforming both the Initial Lexical Model (0.3082) and the Contextual Reddit Model (0.2974).

Finally, we tested to if the Big Five structure would emerge from the IPIP items themselves using machine learning methods. We performed K-Means clustering on the IPIP items themselves (k=5) to see if they would naturally group along theoretical lines. This analysis showed four of the Big Five dimensions were detected in the IPIP by our methods. For example, one data-driven IPIP cluster was most semantically similar to the Big Five trait of Neuroticism (Similarity: 0.5630) and was represented by items such as "Get upset by unpleasant thoughts that come into my mind.". Another cluster mapped clearly to Openness (Similarity: 0.5786) and included items like "Love to think up new ways of doing things.". Surprisingly, none of the IPIP clusters matched the trait of Extraversion for the Big Five model we used, which was also the least present trait in the top 10 subreddits. Taken together, these validation results show that while our bottom-up models identify psychologically relevant themes, the top-down Big Five framework provides the most coherent and numerically superior structure for organizing personality-descriptive language.

**Discussion**

This study addresses a central question at the intersection of psychology and artificial intelligence: is the statistical structure of trait language, as captured by embeddings in machine learning methods, sufficient to produce a meaningful, descriptive model of personality? Our findings reveal a significant mismatch between these purely data-driven semantic structures and the established, human-curated Big Five framework. Specifically, we demonstrate that while a bottom-up machine learning approach identifies coherent semantic clusters, these clusters fail to provide the descriptive power and discriminant validity of the theory-driven Big Five model when applied to a large, naturalistic dataset. This work contributes to the literature in three main ways: by testing the ecological validity of a purely lexical model, by revealing a critical semantic gap in how machines process personality language, and by identifying an anomaly in the representation of Extraversion in these models.

In this study, we used language model embeddings to test multiple personality frameworks, showing a significant mismatch between data-driven semantic structures and established human-created psychological constructs. Our work highlights this mismatch in three main ways. First, we demonstrate that a data-driven approach does not show superiority over the Big Five in discriminant validity for subreddits. The established Big Five model successfully captured nuanced linguistic differences between communities, whereas our bottom-up Lexical and Contextual models performed poorly, failing to provide meaningful distinctions. This suggests that while some studies (van der Linden et al., 2025)have lauded the use of machine learning tools in personality psychology, their ability to generate a useful descriptive model from the bottom up is poorer than applying a top-down, theory-driven framework. Second, our validation analysis showed that even when applying machine learning to a source with a previously established factor structure—the full IPIP-NEO inventory (Goldberg & Saucier, 2008)—the way computerized methods clustered the items was fundamentally different from the structure derived from traditional survey methods; notably, our analysis failed to recover the trait of Extraversion. Third, our study showed that the data-driven analysis of the Reddit corpus did have some positive features, successfully identifying broad, general themes of discourse on the site, such as a "general description" cluster and a "negative affect" cluster, even as it failed to produce a multifaceted personality model. These broad, evaluative clusters may be a reflection of the General Factor of Personality, which recent work has shown is the largest and most dominant factor to emerge from a bottom-up, computational analysis of trait words in natural language using LLMs (van der Linden et al., 2025).

These findings contribute to an evolving discussion on the role of machine learning in psychological science. Our results serve as a direct test of the "deep lexical hypothesis" approach, which had previously recovered a stable, three-factor personality structure directly from the semantic relations between trait words (Cutler & Condon, 2023). While that work demonstrated that a psycho-lexical structure is embedded in language, our findings reveal that this purely semantic structure lacks descriptive utility in a naturalistic context. This failure to generate a useful model from the bottom up creates a stark contrast with recent research showing that large language models can produce remarkably consistent and valid personality profiles when responding to human-created psychometric inventories (Pellert et al., 2024). This suggests a fundamental "semantic gap": LLMs can convincingly mimic human personality traits within a pre-defined theoretical structure, but may struggle to derive that structure in a bottom-up method. This aligns with findings that semantic embeddings are powerful tools for analyzing and resolving "jingle jangle" overlap issues within existing human-curated scales(Wulff & Mata, 2025), further highlighting that these models excel at working with established constructs, rather than generating them. Furthermore, our surprising finding that Extraversion was poorly recovered or conflated with Agreeableness may reflect the models' amplification of previously demonstrated perceptual overlaps of those high in either trait (Rapp et al., 2019). Specifically, previous research has made the case that highly agreeable and extraverted individuals are considered to be more sociable and gregarious (McCrae & Costa, 2003). For example, a network analysis of 268 facet scales from 13 psychometric inventories created a facet atlas; this analysis found that over half of all personality facets are "blended", representing multiple Big Five domains (Schwaba et al., 2020). Crucially, that study noted that blends involving Agreeableness were the most common, challenging simple hierarchical models and providing a data-driven reason why a model might struggle to separate the two traits. Viewed together, our findings help show the boundaries of current computational methods, suggesting they are more effective as tools for validating and refining existing theories of human behavior than creating new ones.

Our findings have significant implications for several fields. For psychology in general, these findings suggest that the lexical hypothesis cannot be reproduced purely through a bottom-up model of lexical similarity. Using psychometric scales is crucial to understanding how people naturally see traits, and therefore, how they use language. In addition, these findings bolster the robustness of the Big Five, as it accurately describes how people use language far above that of other methods. The consistent failure of our models to recover Extraversion, however, presents a significant puzzle. Rather than suggesting the trait is non-existent, our results should be interpreted in the context of how machine learning methods consider how Extraversion is encoded in language. It may be that its structure is less semantically obvious than other traits (especially to machine learning methods) and more dependent on contextual human interpretation. There is some research that may clarify why this is the case. Some research has argued that Extraversion appears to correlate with Agreeableness in cross cultural research, being conflated with Agreeableness—which matches our findings (Laajaj et al., 2019). Future researchers should examine how language related to Extraversion is processed by these machine learning tools, and if any methods result in greater separation between Agreeableness and Extraversion.

Our findings also have significant implications for the field of machine learning and how text is processed by these algorithms. There is sizeable interest in using machine learning tools (like LLMs) to process text and extract meaning from large corpuses, including through qualitative research for a number of purposes (Castellanos et al., 2025; Hayes, 2025). However, if these algorithms do not understand and cluster text the way humans do, there is a possibility that they will do poorly at psychologically sensitive tasks. This may be a particularly problematic issue if the LLM is used without a strong theoretical framework or human oversight, as it risks generating interpretations that are statistically plausible (like in our study) but psychologically meaningless.

We had some major limitations in our methods. We focused on mpnet-personality, an embedding model that focused specifically on personality (Wulff, 2025), but it may be that the embedding method is inadequate or resulted in an unnatural advantage of the Big Five because of the way it was created. In addition, we chose to use K-Means clustering, but personality measures may not cluster in a way that K-means can detect. We did test other models (mpnet general, BERT), and other clustering methods, but there was little difference between them, so we chose to keep the results as they were, especially to compare to other research (Feraco & Toffalini, 2025; Guenole et al., 2025; Wulff & Mata, 2025). This does not preclude the possibility that entirely new methods are needed.

The scope of our data also presents limitations. Our analysis was restricted to English-language text from Reddit, and the surprising scarcity of Extraversion-related language could reflect the platform's user base rather than a universal feature of language. In addition, we did not use data on how people actually respond to the IPIP items, which can be converted into actual overlap. This may improve the model, but the data on IPIP may not be able to be processed well by these machine learning models. Rotation decisions especially may lead to more problems; it may be that orthogonal rotation (where it is assumed factors are unrelated) vs oblique rotations (where it is assumed they may be correlated) are not very well handled by machine learning algorithms.

Finally, our use (and operationalization) of the Big Five as our chosen framework has its own constraints. By focusing on the five broad domains, our top-down model did not account for the more granular facet level, which may be a more appropriate level of analysis. Future research could also incorporate alternative structural models, such as the six-factor HEXACO model, which may offer a better fit. Future researchers may also wish to create a top-down model that focuses on facets rather than the broad Big Five traits alone.

Conclusion

Overall, our work suggests that while language model embeddings can identify coherent semantic patterns, they may not be able to fully capture the nuanced structure of human personality as understood through decades of psychometric research, especially through more robust psychometric methods like survey methods. Attempting to derive the Big Five from personality inventories and Reddit data did have some issues but was far more coherent than bottom-up approaches. This "semantic gap" reveals that embedding models, while powerful, do not "understand" personality in the same way humans do, nor how certain qualities of people (traits) tend to co-occur. This is a critical area for future research at the intersection of psychology and AI, focusing on bridging the gap between data-driven patterns and the way humans perceive patterns of behavior.

**Method**

Our study was designed to re-examine the structure of personality by applying modern computational methods to the foundational lexical hypothesis. The lexical hypothesis posits that the most significant individual differences are encoded within natural language. Historically, researchers like Allport and Odbert (Allport & Odbert, 1936), and later Cattell (Cattell, 1945), sought to uncover a taxonomy of personality by systematically reducing the vast number of trait-descriptive words into a core set of factors. This foundational work, however, relied on manual, subjective judgments to collapse semantically overlapping terms, a process that may have introduced biases that were perpetuated in subsequent research. Our methodology follows the theoretical path of this early work, but replaces subjective curation with objective, data-driven clustering of semantic embeddings to derive structure directly from the language itself. The purpose of this was to reduce the subjectivity, but also increase the transparency, and result in refinements that others can adjust depending on the type of language.

*Procedure:*

*Data sources and preprocessing:*

Trait-Descriptive Adjective Corpus. To begin from a similar starting point as the original psycholexical researchers, we used the dataset of 2,818 trait-descriptive adjectives compiled by Condon, Coughlin, and Weston (Condon et al., 2022). This comprehensive list was created from the foundational psycholexical research, incorporating terms studied by (Allport & Odbert, 1936), (Norman, 1967), and (Goldberg, 1982). This list was processed to ensure all terms were unique and lowercase for the analysis.

Online Discussion Corpus. To analyze language in a naturalistic context, we used a corpus of online discussions from the fddemarco/pushshift-reddit dataset (Demarco, 2023). We examined comments from 16 distinct English-language communities (subreddits) selected to represent a wide range of social, personal, and informational topics. Our analysis was conducted by processing up to one million comments from this stream and identifying all occurrences of adjectives from our primary corpus. for validation.

Validation Corpus. To validate all three models against established psychometric constructs, we used the 300 items from the International Personality Item Pool representation of the NEO PI-R (Goldberg & Saucier, 2008) This public-domain inventory is designed to measure the Big Five personality traits.

*Computational analysis*

All textual data were converted into 768-dimensional vector representations using the dwulff/mpnet-personality model, a sentence transformer fine-tuended on personality related text. Before clustering, we scaled the embeddings using the StandardScaler function from the scikit-learning library to make sure each feature had a mean of 0 and standard deviation of 1. We then applied K-Means clustering from the scikit-learn library to partition these embeddings. Cluster quality was assessed internally by calculating a cohesion score, defined as the mean cosine similarity of member words to their cluster centroid.

*Model construction:*

We constructed three distinct models of personality structure. The first two were data driven, while the third was "top down". For these data-driven models, the number of clusters was determined by performing a silhouette analysis for k-values from 2 to 10, which indicated that k=6 provided a balance of cluster separation and cohesion. The first data driven model, which we call the Lexical Model, applied K-Means clustering (k=6) to the embeddings of the Condon et al (2022) list of 2,818 adjectives. The second data driven model, which we call the Contextual Reddit Model, applied K-Means clustering (k=6) to the embeddings of the 1,945 adjectives in the Reddit corpus. The last model was a theoretical Big Five Model, constructed by creating a centroid vector to each of the Big Five traits. This was accomplished by averaging the embeddings of the adjectives shown to be markers of each trait, both theoretically and practically, in the IPIP (Goldberg, 1992).

*Model comparison:*

Our first analysis was a direct comparison of the descriptive power of the three models. To achieve this, we first mapped every adjective mention found in the Reddit corpus back to its corresponding group in each of the three models (the 6 initial lexical clusters, the 6 contextual Reddit clusters, and the 5 theoretical Big Five traits). We then generated a linguistic profile for each of the top 10 most active subreddits by calculating the distribution of these group assignments as a percentage of total adjective use within that community. The model demonstrating the most nuanced and interpretable differentiation (that was semantically cohesive) between the subreddit profiles was deemed to have the superior descriptive utility.

**Data availability**

The data of this study are available as an attachment.

**Code availability**

All code is available on Github at https://github.com/AUBee19/MLpersonality. Analysis was largely completed using the Python programing language (v.3.13)

**Acknowledgements**

Author 1's position was supported by Dr Elias Aboujaoude at Cedars-Sinai prior to this manuscript's completion. Dr Ayoub Bouguettaya was also trained by Dr Kyle Coleman on machine learning methods, and provided early feedback on the code as part of Cedars-Sinai's AI Campus initiative. Both are acknowledged for their valuable feedback on a very early version of this manuscript.

**Funding**

None to declare.

**Ethics declarations**

Competing interests:

Author 1 will begin employment at Meta on September 22, 2025. This manuscript was completed and submitted prior to the start of this employment. Meta had no role in the conception, design, data collection, analysis, or writing of this manuscript. All other authors declare no competing interests.

# References

Allport, G. W. (1927). Concepts of trait and personality. *Psychological Bulletin*, *24*(5), 284.

Allport, G. W., & Odbert, H. S. (1936). Trait-names: A psycho-lexical study. *Psychological monographs*, *47*(1), i.

Block, J. (2010). The five-factor framing of personality and beyond: Some ruminations. *Psychological Inquiry*, *21*(1), 2-25.

Castellanos, A., Jiang, H., Gomes, P., Vander Meer, D., & Castillo, A. (2025). Large Language Models for Thematic Summarization in Qualitative Health Care Research: Comparative Analysis of Model and Human Performance. *JMIR AI*, *4*, e64447. https://doi.org/10.2196/64447

Cattell, H. E. (1996). The original big five: A historical perspective. *European Review of Applied Psychology/Revue Européenne de Psychologie Appliquée*.

Cattell, R. B. (1945). The description of personality: Principles and findings in a factor analysis. *The American journal of psychology*, *58*(1), 69-90.

Condon, D. M., Coughlin, J., & Weston, S. J. (2022). Personality trait descriptors: 2,818 trait descriptive adjectives characterized by familiarity, frequency of use, and prior use in psycholexical research. *Journal of Open Psychology Data*, *10*(1).

Costa Jr, P. T. (1996). Work and personality: Use of the NEO-PI-R in industrial/organisational psychology. *Applied Psychology*, *45*(3), 225-241.

Costa, P. T., & McCrae, R. R. (1992). Normal personality assessment in clinical practice: The NEO Personality Inventory. *Psychological assessment*, *4*(1), 5.

Costa, P. T., & McCrae, R. R. (2008). The revised neo personality inventory (neo-pi-r). *The SAGE handbook of personality theory and assessment*, *2*(2), 179-198.

Cutler, A., & Condon, D. M. (2023). Deep lexical hypothesis: Identifying personality structure in natural language. *J Pers Soc Psychol*, *125*(1), 173-197. https://doi.org/10.1037/pspp0000443

Decuyper, M., De Fruyt, F., & Buschman, J. (2008). A five-factor model perspective on psychopathy and comorbid Axis-II disorders in a forensic-psychiatric sample. *Int J Law Psychiatry*, *31*(5), 394-406. https://doi.org/10.1016/j.ijlp.2008.08.008

Demarco, F. (2023). *Pushshift-reddit* https://huggingface.co/datasets/fddemarco/pushshift-reddit

Du, C. D., Fu, K. C., Wen, B. C., Sun, Y., Peng, J., Wei, W., Gao, Y., Wang, S. P., Zhang, C. C., Li, J. P., Qiu, S., Chang, L., & He, H. G. (2025). Human-like object concept representations emerge naturally in multimodal large language models. *Nature Machine Intelligence*, *7*(6), 1-16. https://doi.org/10.1038/s42256-025-01049-z

Feraco, T., & Toffalini, E. (2025). SEMbeddings: how to evaluate model misfit before data collection using large-language models. *Frontiers in Psychology*, *15*, 1433339.

Gigerenzer, G. (2024). Psychological AI: Designing Algorithms Informed by Human Psychology. *Perspect Psychol Sci*, *19*(5), 839-848. https://doi.org/10.1177/17456916231180597

Goldberg, L., & Saucier, G. (2008). The Eugene-Springfield community sample: Information available from the research participants. *Oregon Research Institute Technical Report*, *48*(1).

Goldberg, L. R. (1982). From Ace to Zombie: Some explorations in the language of personality. *Advances in personality assessment*, *1*, 203-234.

Goldberg, L. R. (1992). The development of markers for the Big-Five factor structure. *Psychological assessment*, *4*(1), 26.

Goldberg, L. R. (1993). The structure of phenotypic personality traits. *Am Psychol*, *48*(1), 26-34. https://doi.org/10.1037//0003-066x.48.1.26

Guenole, N., D'Urso, E. D., Samo, A., Sun, T., & Haslbeck, J. M. B. (2025). Enhancing Scale Development: Pseudo Factor Analysis of Language Embedding Similarity Matrices. https://doi.org/10.31234/osf.io/vf3se_v2

Hayes, A. S. (2025). "Conversing" With Qualitative Data: Enhancing Qualitative Research Through Large Language Models (LLMs). *International Journal of Qualitative Methods*, *24*, 16094069251322346.

Laajaj, R., Macours, K., Pinzon Hernandez, D. A., Arias, O., Gosling, S. D., Potter, J., Rubio-Codina, M., & Vakis, R. (2019). Challenges to capture the big five personality traits in non-WEIRD populations. *Sci Adv*, *5*(7), eaaw5226. https://doi.org/10.1126/sciadv.aaw5226

McCrae, R. R., & Costa Jr, P. T. (1997). Personality trait structure as a human universal. *American psychologist*, *52*(5), 509.

McCrae, R. R., & Costa, P. T. (2003). *Personality in adulthood: A five-factor theory perspective*. Guilford press.

Norman, W. T. (1967). On estimating psychological relationships: social desirability and self-report. *Psychol Bull*, *67*(4), 273-293. https://doi.org/10.1037/h0024414

Pellert, M., Lechner, C. M., Wagner, C., Rammstedt, B., & Strohmaier, M. (2024). Ai psychometrics: Assessing the psychological profiles of large language models through psychometric inventories. *Perspectives on Psychological Science*, *19*(5), 808-826.

Rapp, C., Ingold, K., & Freitag, M. (2019). Personalized networks? How the Big Five personality traits influence the structure of egocentric networks. *Social science research*, *77*, 148-160.

Schwaba, T., Rhemtulla, M., Hopwood, C. J., & Bleidorn, W. (2020). A facet atlas: Visualizing networks that describe the blends, cores, and peripheries of personality structure. *PLOS ONE*, *15*(7), e0236893. https://doi.org/10.1371/journal.pone.0236893

van der Linden, D., Cutler, A., Van der Linden, P. A., & Dunkel, C. S. (2025). The general factor of personality (GFP) in natural language: A deep learning approach. *Journal of Research in Personality*, 104635.

Wulff, D. U. (2025). mpnet-personality. In dwulff/mpnet-personality (Ed.): Hugging Face.

Wulff, D. U., & Mata, R. (2025). Semantic embeddings reveal and address taxonomic incommensurability in psychological measurement. *Nat Hum Behav*, *9*(5), 944-954. https://doi.org/10.1038/s41562-024-02089-y